\documentclass[10pt,twocolumn,letterpaper]{article}

\usepackage{iccv}
\usepackage{times}
\usepackage{epsfig}
\usepackage{graphicx}
\usepackage{amsmath}
\usepackage{amssymb}

\usepackage{booktabs}
\usepackage{makecell}
\usepackage{multirow}
\usepackage{tabularx}
\usepackage{bbm}
\usepackage{caption} 
\usepackage{subcaption}

\newcolumntype{L}[1]{>{\raggedright\let\newline\\\arraybackslash\hspace{0pt}}m{#1}}
\newcolumntype{C}[1]{>{\centering\let\newline\\\arraybackslash\hspace{0pt}}m{#1}}
\newcolumntype{R}[1]{>{\raggedleft\let\newline\\\arraybackslash\hspace{0pt}}m{#1}}
\newcolumntype{Y}{>{\centering\arraybackslash}X}

\makeatletter
\renewcommand{\paragraph}{%
  \@startsection{paragraph}{4}%
  {\z@}{1.1ex \@plus 1ex \@minus .2ex}{-1em}%
  {\normalfont\normalsize\bfseries}%
}
\makeatother

\usepackage[pagebackref=true,breaklinks=true,letterpaper=true,colorlinks,bookmarks=false]{hyperref}

\iccvfinalcopy %

\def\iccvPaperID{8134} %

\ificcvfinal\fi

\begin{document}

\title{iMAP: Implicit Mapping and Positioning in Real-Time}

\author{Edgar Sucar$^{1}$\quad Shikun Liu$^{1}$\quad Joseph Ortiz$^{2}$ \quad Andrew J. Davison$^{1}$ \\
$^{1}$Dyson Robotics Lab, Imperial College London\\
$^{2}$Robot Vision Lab, Imperial College London\\
{\tt\small \{e.sucar18, shikun.liu17, j.ortiz, a.davison\}@imperial.ac.uk}
}

\maketitle

\begin{abstract}
We show for the first time that a multilayer perceptron (MLP) can serve as the only scene representation in a real-time SLAM system for a handheld RGB-D camera. Our network is trained in live operation without prior data, building a dense, scene-specific implicit 3D model of occupancy and colour which is also immediately used for tracking. 

Achieving real-time SLAM via continual training of a neural network against a live image stream requires significant innovation. Our iMAP algorithm uses a keyframe structure and multi-processing computation flow, with dynamic information-guided pixel sampling for speed, with tracking at 10 Hz and global map updating at 2 Hz. The advantages of an implicit MLP over standard dense SLAM techniques include efficient geometry representation with automatic detail control and smooth, plausible filling-in of unobserved regions such as the back surfaces of objects.

\end{abstract}

\section{Introduction}
A real-time Simultaneous Localisation and Mapping (SLAM) system for an intelligent embodied device must incrementally build a representation of the 3D world, to enable both localisation and scene understanding. The ideal representation should precisely encode geometry, but also be {\em efficient},
with the memory capacity available used adaptively in response to scene size and complexity;
{\em predictive}, able to plausibly estimate the shape of regions not directly observed;
and {\em flexible}, not needing a large amount of training data or manual adjustment to run in a new scenario.

\begin{figure}[t]
\begin{center}
   \includegraphics[width=\linewidth]{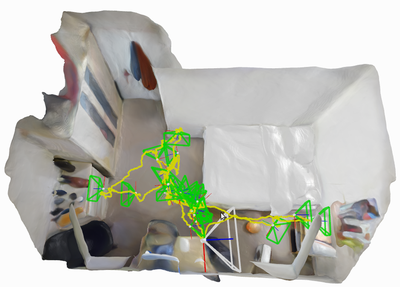}
\end{center}
\caption{Room reconstruction from real-time iMAP with an Azure Kinect RGB-D camera, showing watertight scene model, camera tracking and automatic keyframe set.}
\label{fig:teaser}
    \vspace{2mm} \hrule
\end{figure}

Implicit neural representations are a promising recent advance in off-line reconstruction, using a multilayer perceptron (MLP) to map a query 3D point to occupancy or colour, and optimising it from scratch to fit a specific scene.  An MLP is a general implicit function approximator, able to represent variable detail with few parameters and without quantisation artifacts.
Even without prior training, the inherent priors present in the network structure allow it to make watertight geometry estimates from partial data, and plausible completion of unobserved regions.

In this paper, we show for the first time that an MLP can be used as the only scene representation in a real-time SLAM system using a hand-held RGB-D camera.
Our randomly-initialised network is trained in {\it live operation} and we do not require any prior training data. 
Our iMAP system is designed with a keyframe structure and multi-processing computation flow reminiscent of PTAM~\cite{Klein:Murray:ISMAR2007}. 
In a tracking process, running at over 10 Hz, we align live RGB-D observations with rendered depth and colour predictions from the MLP scene map. In parallel, a mapping process selects and maintains a set of historic keyframes whose viewpoints span the scene, and uses these to continually train and improve the MLP, while jointly optimising the keyframe poses.

In both tracking and mapping, we dynamically sample the most informative RGB-D pixels to reduce geometric uncertainty, achieving real-time speed.
Our system runs in Python, and all optimisation is via a standard PyTorch framework \cite{paszke2019pytorch} on a single desktop CPU/GPU system.

By casting SLAM as a continual learning problem, we achieve a representation which can represent scenes efficiently with continuous and adaptive resolution, and with a remarkable ability to smoothly interpolate to achieve complete, watertight reconstruction (Fig. \ref{fig:teaser}). With around 10 - 20 keyframes, and an MLP with only 1 MB of parameters, we can accurately map whole rooms. Our scene representation has no fixed resolution; the distribution of keyframes automatically achieves efficient multi-scale mapping.

We demonstrate our system on a wide variety of real-world sequences and do exhaustive evaluation and ablative analysis on 8 scenes from the room-scale Replica Dataset \cite{Straub:etal:ARXIV2019}. We show that iMAP can make a more \textbf{complete scene reconstruction} than standard dense SLAM systems with \textbf{significantly smaller memory footprint}. We show competitive tracking performance on the TUM RGB-D dataset \cite{Sturm:etal:IROS2012} against state-of-the-art SLAM systems.

To summarise, the key contributions of the paper are:
\begin{itemize}
    \item The first dense real-time SLAM system that uses an implicit neural scene representation and is capable of jointly optimising a full 3D map and camera poses.
    \item The ability to {\it incrementally} train an implicit scene network in real-time, enabled by automated keyframe selection and loss guided sparse active sampling.
    \item A parallel implementation (fully in PyTorch \cite{paszke2019pytorch} with multi-processing) of our presented SLAM formulation which works online with a hand-held RGB-D camera. %
\end{itemize}

\section{Related Work}
\paragraph{Visual SLAM Systems}

Real-time visual SLAM systems for modelling environments are often built in a layered manner, where a sparse representation is used for localisation and more detailed geometry or semantics is layered on top. However, here we work in the `dense SLAM' paradigm pioneered in~\cite{Newcombe:etal:ICCV2011,Newcombe:etal:ISMAR2011} where a {\it unified} dense scene representation is also the basis for camera tracking. Dense representations avoid arbitrary abstractions such as keypoints, enable tracking and relocalisation in robust invariant ways, and have long-term appeal as sensor-agnostic, unified, complete representations of spaces.

Some approaches in dense SLAM explicitly represent surfaces~\cite{Keller:etal:3DV2013,Whelan:etal:RSS2015}, but  direct representation of volume is desirable to enable a full range of applications such as planning. Standard representations for volume using occupancy or signed distance functions are very expensive in terms of memory if a fixed resolution is used~\cite{Newcombe:etal:ISMAR2011}. Hierarchical approaches~\cite{Dai:etal:ACMTOG2017,Vespa:etal:RAL2018,schops2019bad} are more efficient, but are complicated to implement and usually offer only a small range of level of detail. In either case, the representations are rather rigid, and not amenable to joint optimisation with camera poses, due to the huge number of parameters they use.

Machine learning can discover low-dimensional embeddings of dense structure which enable efficient, jointly optimisable representation.  CodeSLAM~\cite{Bloesch:etal:CVPR2018} is one example, but using a depth-map view representation rather than full volumetric 3D. Learning techniques have also been used to improve dense reconstruction but require an existing scan \cite{dai2020sg} or previous training data \cite{Peng2020ECCV, weder2020routedfusion, chabra2020deep}.

\paragraph{Implicit Scene Representation with MLPs} 

Scene representation and graphics have seen much recent progress on using implicit MLP neural models for object reconstruction \cite{park2019deepsdf,mescheder2019occupancy}, object compression \cite{tang2020deep} novel view synthesis \cite{Mildenhall:etal:ECCV2020}, and scene completion \cite{Sitzmann:etal:NIPS2020,chibane2020neural}. Two recent papers \cite{wang2021nerf, yen2020inerf} have also explored camera pose optimisation. But so far these methods have been considered as an offline tool, with computational requirements on the order of hours, days or weeks.
We show that when depth images are available, and when guided sparse sampling is used for rendering and training, these methods are suitable for real-time SLAM.

\paragraph{Continual Learning} By using a single MLP as a master scene model, we pose real-time SLAM as {\it online continual learning}. An effective continual learning system should demonstrate both plasticity (the ability to acquire new knowledge) and stability (preserving old knowledge)~\cite{rolnick2019experience,grossberg1982does}. Catastrophic forgetting is a well-known property of neural networks, and is a failure of stability, where new experiences overwrite memories.

One line of work on alleviating catastrophic forgetting has focused on protecting representations against new data using relative weighting~\cite{kirkpatrick2017overcoming}.  This is reminiscent of classic filtering approaches in SLAM such as the EKF~\cite{Smith:Cheeseman:IJRR1986} and is worth future investigation. Approaches which freeze \cite{rusu2016progressive} or consolidate \cite{schwarz2018progress} sub-networks after training on each individual task are perhaps too simple and discrete for SLAM.

Instead, we direct our attention towards the {\it replay-}based approach to continual learning, where previous knowledge is stored either directly in a buffer~\cite{maltoni2019continuous,rolnick2019experience}, or compressed in a generative model~\cite{lesort2019generative,shin2017continual}. We use a straightforward method where keyframes are automatically selected to store and compress past memories. We use loss-guided random sampling of these keyframes in our continually running map update process to periodically replay and strengthen previously-observed scene regions, while continuing to add information via new keyframes. In SLAM terms, this approach is similar to that pioneered by PTAM~\cite{Klein:Murray:ISMAR2007}, where a historic keyframe set and repeated global bundle adjustment serve as a long-term scene representation.

\section{iMAP: A Real-Time Implicit SLAM System}

\subsection{System Overview}
\begin{figure}[t!]
   \centering
   \includegraphics[width=\linewidth]{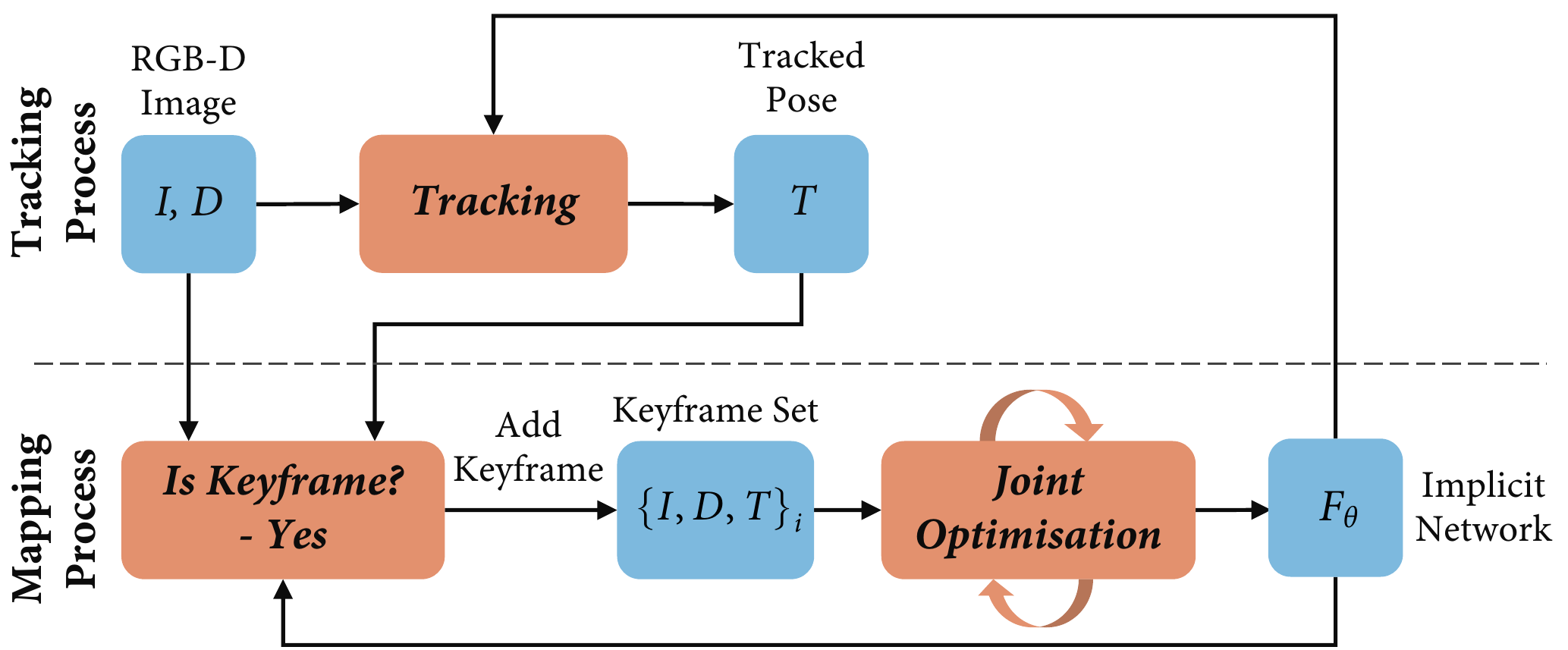}
\caption{iMAP system pipeline.}
\label{fig:slam_diagram}
    \vspace{2mm} \hrule
    \vspace{-2mm}
\end{figure}

Figure \ref{fig:slam_diagram} overviews how iMAP works.
A 3D volumetric map is represented using a fully-connected neural network $F_{\theta}$ that maps a 3D coordinate to colour and volume density (Section \ref{sec:network}). Given a camera pose, we can render the colour and depth of a pixel by accumulating network queries from samples in a back-projected ray (Section \ref{sec:render}). 

We map a scene from depth and colour video by incrementally optimising the network weights and camera poses with respect to a sparse set of actively sampled measurements (Section \ref{sec:active}). Two processes run concurrently: \textit{tracking} (Section \ref{sec:track}), which optimises the pose from the current frame with respect to the locked network; and \textit{mapping} (Section \ref{sec:joint}), which jointly optimises the network and the camera poses of selected keyframes, incrementally chosen based on information gain (Section \ref{sec:keyframe}). 

\subsection{Implicit Scene Neural Network}
\label{sec:network}

Following the network architecture in NeRF \cite{Mildenhall:etal:ECCV2020}, we use an MLP with 4 hidden layers of feature size $256$, and two output heads that map a 3D coordinate $\textbf{p} = (x,y,z)$ to a colour and volume density value: $F_{\theta}(\textbf{p}) = (\textbf{c}, \rho)$. Unlike NeRF, we do not take into account viewing directions as we are not interested in modelling specularities.

We apply the Gaussian positional embedding proposed in Fourier Feature Networks \cite{Tancik:etal:NIPS2020} to lift the input 3D coordinate into $n$-dimensional space: $\sin(\textbf{B}\textbf{p})$, with $\textbf{B}$ an $[n \times 3]$ matrix sampled from a normal distribution with standard deviation $\sigma$. This embedding serves as input to the MLP and is also concatenated to the second activation layer of the network. Taking inspiration from SIREN \cite{Sitzmann:etal:NIPS2020}, we allow optimisation of the embedding matrix $\textbf{B}$, implemented as a single fully-connected layer with sine activation.

\subsection{Depth and Colour Rendering}
\label{sec:render}

Our new differentiable rendering engine, inspired by NeRF \cite{Mildenhall:etal:ECCV2020} and NodeSLAM \cite{Sucar:etal:3DV2020},
queries the scene network to obtain depth and colour images from a given view.

Given a camera pose $T_{WC}$ and a pixel coordinate $[u, v]$, we first back-project a normalised viewing direction and transform it into world coordinates: $\textbf{r} = T_{WC} K^{-1}[u, v]$, with the camera intrinsics matrix $K$ . We take a set of $N$ samples along the ray $\textbf{p}_i = d_i \textbf{r}$ with corresponding depth values $\{d_1, \cdots, d_N\}$, and query the network for a colour and volume density $(\textbf{c}_i, \rho_i) = F_{\theta} (\textbf{p}_i)$. We follow the stratified and hierarchical volume sampling strategies of NeRF.

Volume density is transformed into an occupancy probability by multiplying by the inter-sample distance $\delta_i = d_{i+1} - d_{i}$ and passing this through activation function $o_i = 1 - \exp(-\rho_i
\delta_i)$. The ray termination probability at each sample can then be calculated as $w_i = o_i\prod_{j=1}^{i-1} (1-o_j)$.
Finally, depth and colour are rendered as the expectations:
\begin{equation}
  \hat{D}[u,v] = \sum_{i=1}^N w_i d_i, \quad \hat{I}[u,v] = \sum_{i=1}^N w_i \textbf{c}_i.
\end{equation}
We can calculate the depth variance along the ray as:
\begin{equation}
\hat{D}_{var}[u,v] = \sum_{i=1}^N w_i (\hat{D}[u,v] - d_i)^2.
\end{equation}

\subsection{Joint optimisation}
\label{sec:joint}

We jointly optimise the implicit scene network parameters $\theta$, and camera poses for a growing set of $W$ keyframes, each of which has associated colour and depth measurements along with an initial pose estimate: $\{I_i, D_i, T_i\}$.

Our rendering function is differentiable with respect to these variables, so we perform iterative optimisation to minimise the geometric and photometric errors for a selected number of rendered pixels $s_i$ in each keyframe.
 
The photometric loss is the L1-norm between the rendered and measured colour values $ e_i^p[u,v] = \left|I_i[u,v] - \hat{I}_i[u,v]\right|$ for $M$ pixel samples:
\begin{equation}
    L_p = \frac{1}{M} \sum_{i=1}^W \sum_{(u,v) \in s_i} e_i^p[u,v].
\end{equation}
The geometric loss measures the depth difference $ e_i^g[u,v] = \left|D_i[u,v] - \hat{D}_i[u,v]\right|$ and uses the depth variance as a normalisation factor, down-weighting the loss in uncertain regions such as object borders:
\begin{equation}
    L_g = \frac{1}{M} \sum_{i=1}^W \sum_{(u,v) \in s_i} \frac{e_i^g[u,v]}{\sqrt{\hat{D}_{var}[u, v]}}.
\end{equation}

We apply the ADAM optimiser \cite{Kingma:etal:ICLR2015} on the weighted sum of both losses, with factor $\lambda_p$ adjusting the importance given to the photometric error:
\begin{equation}
    \min \limits_{\theta, \{T_i\}} (L_g + \lambda_p L_p)~.
\end{equation}

\paragraph{Camera Tracking}
\label{sec:track}

In online SLAM, close to frame-rate camera tracking is important, as optimisation of smaller displacements is more robust. We run a parallel tracking process that continuously optimises the pose of the latest frame with respect to the fixed scene network at a much higher frame rate than joint optimisation while using the same loss and optimiser. The tracked pose initialisation is refined in the mapping process for selected keyframes.

\subsection{Keyframe Selection}
\label{sec:keyframe}

Jointly optimising the network parameters and camera poses using all images from a video stream is not computationally feasible. However, since there is huge redundancy in video images, we may represent a scene with a sparse set of representative keyframes, incrementally selected based on {\it information gain}. The first frame is always selected to initialise the network and fix the world coordinate frame. Every time a new keyframe is added, we lock a copy of our network to represent a snapshot of our 3D map at that point in time. Subsequent frames are checked against this copy and are selected if they see a significantly new region.

For this, we render a uniform set of pixel samples $s$ and calculate the proportion $P$ with a normalised depth error smaller than threshold $t_D = 0.1$, to measure the fraction of the frame already explained by our map snapshot:
\begin{equation}
    P = \frac{1}{|s|}  \sum_{(u,v) \in s} \mathbbm{1} \left(\frac{\left|D[u,v] - \hat{D}[u,v]\right|}{D[u,v]} < t_D \right).
\end{equation}

When this proportion falls under a threshold $P<t_P$ (we set $t_P=0.65$), this frame is added to the keyframe set. The normalised depth error produces adaptive keyframe selection, requiring higher precision, and therefore more closely spaced keyframes, when the camera is closer to objects.

Every frame received in the mapping process is used in joint optimisation for a few iterations (between 10 and 20), so our keyframe set is always composed of the selected set along with the continuously changing latest frame. 

\subsection{Active Sampling}
\label{sec:active}

\paragraph{Image Active Sampling} 
Rendering and optimising all image pixels would be expensive in computation and memory. We take advantage of image regularity to render and optimise only a very sparse set of random pixels (200 per image) at each iteration. Further, we use the render loss to guide active sampling in informative areas with higher detail or where reconstruction is not yet precise. 

\begin{figure}[t]
    \centering
    \includegraphics[width=0.54\linewidth]{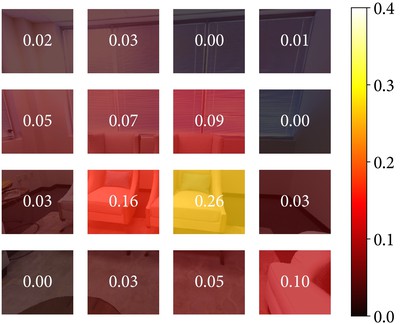}
    \hfill
    \includegraphics[width=0.45\linewidth]{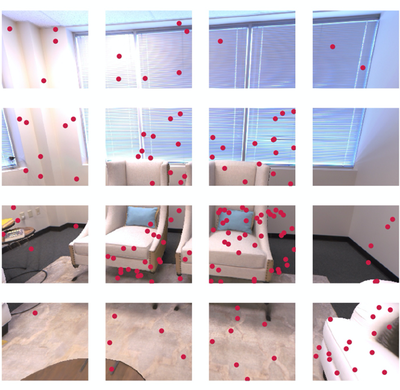}
    \caption{Image Active Sampling. Left: a loss distribution is calculated across an image grid using the geometric loss from a set of uniform samples. Right: active samples are further allocated proportional to the loss distribution.}
    \label{fig:active}
    \vspace{2mm} \hrule 
    \vspace{-0.2cm}
\end{figure}

Each joint optimisation iteration is divided into two stages. First, we sample a set $s_i$ of pixels, uniformly distributed across each of the keyframe's depth and colour images. These pixels are used to update the network and camera poses, and to calculate the loss statistics. For this, we divide each image into an $[8 \times 8]$ grid, and calculate the average loss inside each square region $R_j$, $j = \{1,2, \cdots,64\}$:
\begin{equation}
    L_i[j] = \frac{1}{|r_{j}|} \sum_{(u,v) \in r_{j}} e_i^g[u,v] + e_i^p[u,v] ,
\end{equation}
where $r_{j} = s_i \cap R_{j}$ are pixels uniformly sampled from $R_{j}$. We normalise these statistics into a probability distribution:
\begin{equation}
    f_i[j] = \frac{L_i[j]}{\sum_{m=1}^{64} L_i[m]}.
\end{equation}
We use this distribution to re-sample a new set of  $n_i \cdot f_i[j]$ uniform samples per region ($n_i$ is the total samples in each keyframe), allocating more samples to regions with high loss. The scene network is updated with the loss from active samples (in camera tracking only uniform sampling is used). Image active sampling is illustrated in Fig. \ref{fig:active}.

\begin{figure}[ht!]
    \centering
    \includegraphics[width=\linewidth]{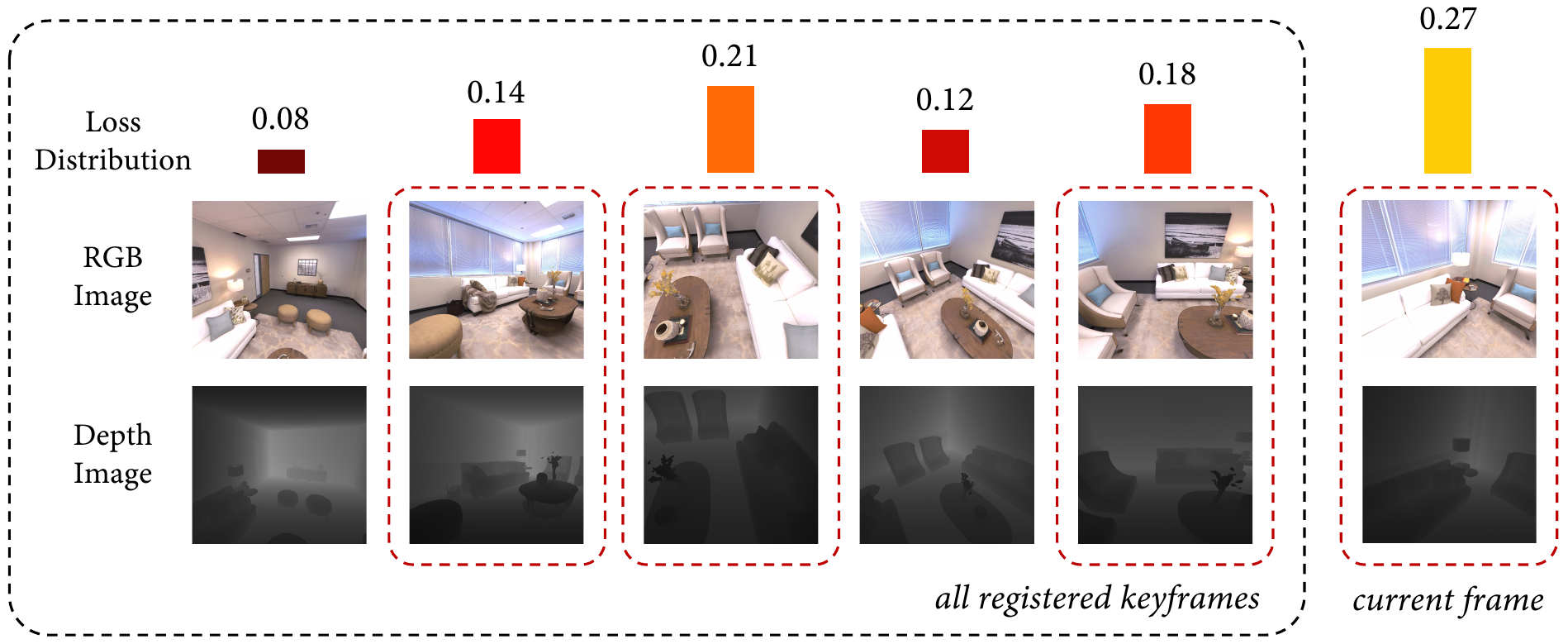}
    \caption{Keyframe Active Sampling. We maintain a loss distribution over the registered keyframes. The distribution is used for sampling a bounded  window of keyframes (red boxes), and for allocating pixel samples in each.}
    \label{fig:keyframe_samp}
    \vspace{2mm} \hrule 
    \vspace{-0.2cm}
\end{figure}

\paragraph{Keyframe Active Sampling}
In iMAP, we continuously optimise our scene map with a set of selected keyframes, serving as a memory bank to avoid network forgetting.  We wish to allocate more samples to keyframes with a higher loss, because they relate to regions which are newly explored, highly detailed, or that the network started to forget.
We follow a process analogous to image active sampling and allocate $n_i$ samples to each keyframe, proportional to the loss distribution across keyframes, See Fig. \ref{fig:keyframe_samp}.

\paragraph{Bounded Keyframe Selection}
Our keyframe set keeps growing as the camera moves to new and unexplored regions. To bound joint optimisation computation, we choose a fixed number (3 in the live system) of keyframes at each iteration, randomly sampled according to the loss distribution. We always include the last keyframe and the current live frame in joint optimisation, to compose a bounded window with $W=5$ constantly changing frames. See Fig. \ref{fig:keyframe_samp}. 

\section{Experimental Results}
Through comprehensive experiments we evaluate iMAP's 3D reconstruction and tracking, and conduct a detailed ablative analysis of design choices on accuracy and speed. Please see our attached video demonstrations.

\subsection{Experimental Setup}

\paragraph{Datasets} 
We experiment on both simulated and real sequences. For reconstruction evaluation we use the Replica dataset \cite{Straub:etal:ARXIV2019},  high quality 3D reconstructions of real room-scale environments, with 5 offices and 3 apartments. For each Replica scene, we render a random trajectory of 2000 RGB-D frames. For raw camera recordings, we capture RGB-D videos using a hand-held Microsoft Azure Kinect on a wide variety of environments, as well as test on the TUM RGB-D dataset \cite{Sturm:etal:IROS2012} to evaluate camera tracking.

\begin{figure}[b]
  \centering
  \includegraphics[width=\linewidth]{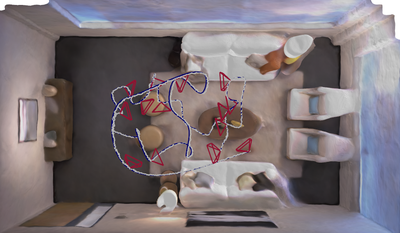}
\caption{Reconstruction and tracking results for Replica {\tt room-0} along with registered keyframes.}
\label{fig:traj}
    \vspace{2mm} \hrule
\end{figure}

\paragraph{Implementation Details} For all experiments we set the following default parameters: keyframe registration threshold $t_P=0.65$, photo-metric loss weighting $\lambda_p=5$, keyframe window size $W=5$, pixel samples $|s_i|=200$, positional embedding size $m=93$ and sigma $\sigma=25$, and $32$ coarse and $12$ fine bins for rendering. 3D point coordinates are normalised by $\frac{1}{10}$ to be close to the $[0, 1]$ range.

In online operation from a hand-held camera, streamed images which arrive between processed frames are dropped. For the experiments presented here every captured frame is processed, running at 10 Hz. We recover mesh reconstructions if needed by querying occupancy values from the network in a uniform voxel grid and then running marching cubes. Meshing is for visualisation and evaluation purposes and does not form part of our SLAM system.

\begin{figure}[t!]
   \centering 
   \includegraphics[width=0.49\linewidth]{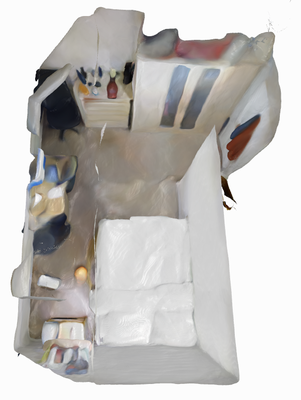}\hfill
   \includegraphics[width=0.49\linewidth]{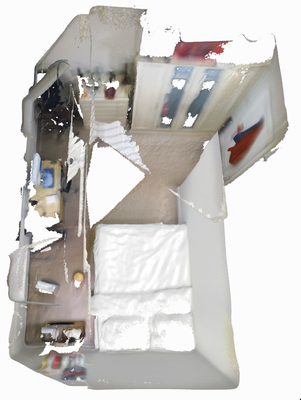}
\caption{iMAP (left) manages to fill in unobserved regions which can be seen as holes in TSDF fusion (right).}
\label{fig:room_comp}
    \vspace{2mm} \hrule
\end{figure}

\subsection{Scene Reconstruction Evaluation}

\paragraph{Metrics} We sample $200,000$ points from both ground-truth and reconstructed meshes, and calculate three quantitative metrics: {\it Accuracy} (cm): the average distance between sampled points from the reconstructed mesh and the nearest ground-truth point; {\it Completion} (cm): the average distance between sampled points from the ground-truth mesh and the nearest reconstructed; and {\it Completion Ratio} ($<$5cm \%): the percentage of points in the reconstructed mesh with {\it Completion} under 5 cm.

\begin{figure*}[ht!]
   \centering
   \includegraphics[width=0.305\linewidth]{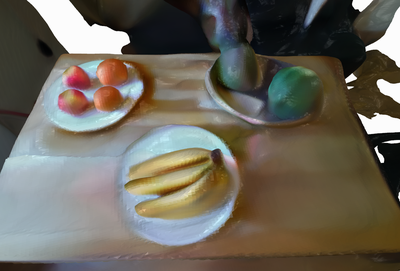}
   \includegraphics[width=0.24\linewidth]{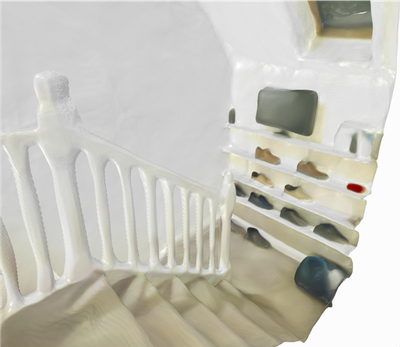}
   \includegraphics[width=0.44\linewidth]{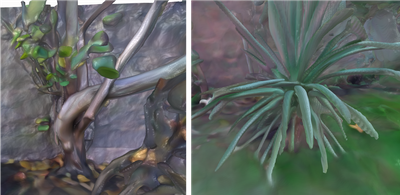}
\caption{Real-time reconstruction results from iMAP in a variety of real world settings.}
\label{fig:add_examples}
    \vspace{2mm} \hrule \vspace{-2mm}
\end{figure*}

The ability to jointly optimise a 3D map along with camera poses gives our system the capacity to build full globally coherent scene reconstructions as seen in Fig. \ref{fig:teaser} and \ref{fig:replica}, and accurate camera tracking as shown in Fig. \ref{fig:traj}. The robustness and versatility of iMAP is demonstrated on a wide variety of real world recordings, through the  reconstructions in Fig. \ref{fig:add_examples} and \ref{fig:real_comp} that show its ability to work at scales from whole rooms to small objects and thin structures. 

We compare scene reconstructions from iMAP with TSDF fusion \cite{Curless:Levoy:SIGGRAPH1996,Newcombe:etal:ISMAR2011}, which is representative of fusion-based dense SLAM methods. To isolate reconstruction, we use the camera tracking produced by iMAP for TSDF fusion. The most significant advantage of our implicit representation is the ability to fill in unobserved regions as shown in Figs. \ref{fig:replica} and \ref{fig:real_comp}. iMAP achieves on average a 4\% higher completion ratio across all 8 Replica scenes as seen in Table \ref{tab:replica}, with an improvement of 11\% in {\tt office-3}. 

\begin{table*}[t!]
  \centering
  \footnotesize
  \setlength{\tabcolsep}{0.5em}
    \begin{tabular}{clccccccccc}
      \toprule
         & & \tt{room-0} & \tt{room-1} & \tt{room-2}  & \tt{office-0} &  \tt{office-1} & \tt{office-2} & \tt{office-3} & \tt{office-4} & Avg. \\
       \midrule
              \multirow{4}{*}{iMAP}   & {\bf \# Keyframes}  
           & 11 & 12 & 12 & 10 & 11 & 10 & 14 & 11 & 13.37 \\
           & {\bf Acc.} [cm] 
           & 3.58 & 3.69 & 4.68 & 5.87 & 3.71 & 4.81 & 4.27 & 4.83 & 4.43 \\
           & {\bf Comp.} [cm] 
           & 5.06 & 4.87 & 5.51 & 6.11 & 5.26  & 5.65 & 5.45 & 6.59 &  \textbf{5.56}\\
           & {\bf Comp. Ratio} [$<$ 5cm \%] 
           & 83.91 & 83.45 & 75.53 & 77.71& 79.64 & 77.22& 77.34 & 77.63 & \textbf{79.06}\\
      \midrule
      \multirow{3}{*}{\makecell{TSDF\\ Fusion}}  & {\bf Acc.} [cm]  
      & 4.21 & 3.08 & 2.88 & 2.70 & 2.66 & 4.27 & 4.07 & 3.70 & \textbf{3.45}\\
      & {\bf Comp.} [cm] 
      & 5.04 & 4.35 & 5.40 & 10.47 & 10.29 & 6.43 & 6.26 & 4.78 & 6.63\\
      & {\bf Comp. Ratio} [$<$ 5cm \%] 
      & 76.90  & 79.87 & 77.79 & 79.60 & 71.93 & 71.66 & 65.87 & 77.11 & 75.09 \\
       \bottomrule
    \end{tabular}%
    \caption{Reconstruction results for 8 indoor Replica scenes. We report the highest reached completion ratio in each scene along with the corresponding accuracy and completion values at that point.}
    \label{tab:replica}
\end{table*}

\begin{figure*}[ht!]
  \centering
  \small
  \setlength{\tabcolsep}{0.1em}
  \begin{tabular}{>{\centering}m{0.06\textwidth} >{\centering}m{0.245\textwidth} >{\centering}m{0.205\textwidth} >{\centering}m{0.215\textwidth} >{\centering\arraybackslash}m{0.255\textwidth}}
    & {\tt room-1} & {\tt room-2} & {\tt office-1} & {\tt office-2} \\
    \makecell{Ground \\ Truth} &  \includegraphics[width=\linewidth]{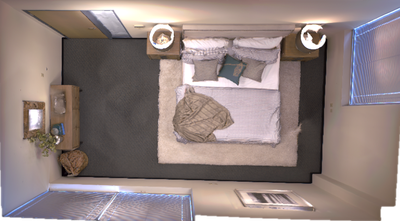} &
    \includegraphics[width=\linewidth]{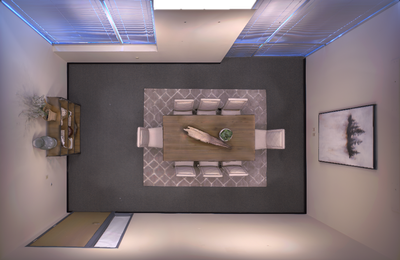} &
    \includegraphics[width=\linewidth]{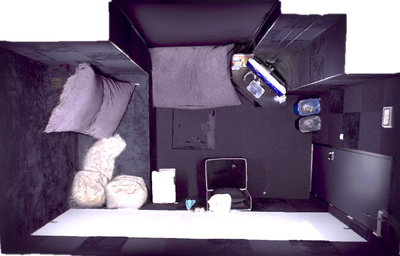} &
    \includegraphics[width=\linewidth]{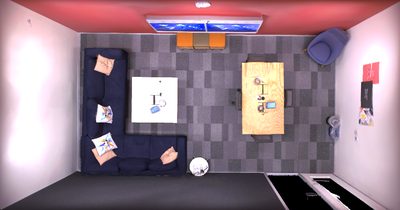} \\
    iMAP &
    \includegraphics[width=\linewidth]{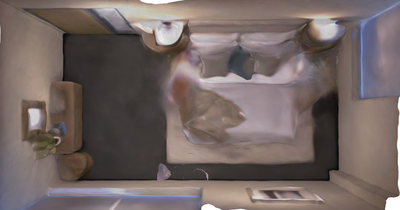} & 
    \includegraphics[width=\linewidth]{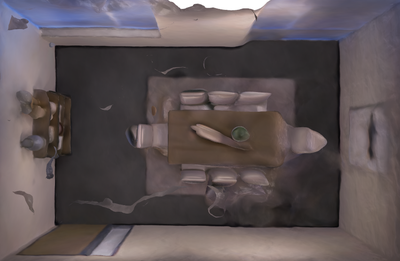} &
    \includegraphics[width=\linewidth]{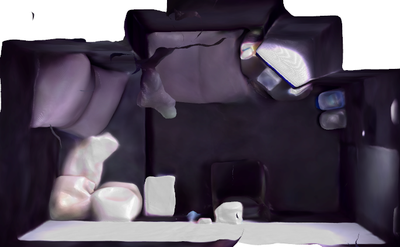} &
    \includegraphics[width=\linewidth]{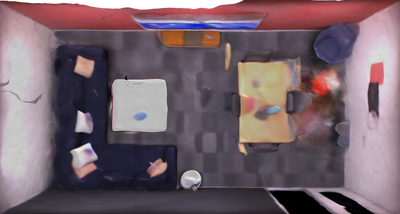} \\
    \makecell{TSDF \\ Fusion} &
    \includegraphics[width=\linewidth]{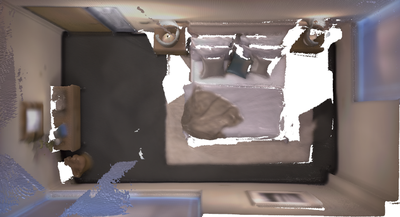} &
    \includegraphics[width=\linewidth]{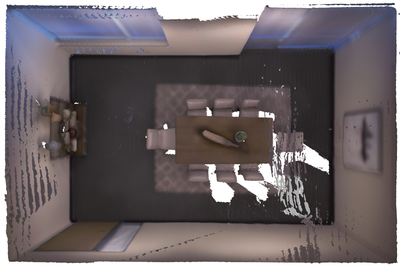} &
    \includegraphics[width=\linewidth]{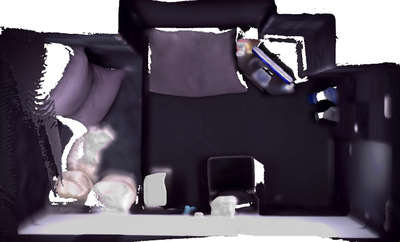} &
    \includegraphics[width=\linewidth]{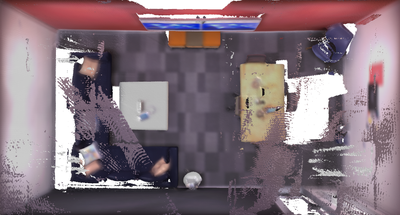} \\
  \end{tabular} 
  \caption{Replica reconstructions, highlighting how iMAP fills in unobserved regions which 
  are white holes in TSDF fusion.}
  \label{fig:replica}
\end{figure*}

\begin{figure*}[ht!]
  \centering
  \small
  \setlength{\tabcolsep}{0.1em}
  \begin{tabular}{>{\centering}m{0.06\textwidth} >{\centering\arraybackslash}m{0.93\textwidth}}
    iMAP &
    \includegraphics[width=0.26\linewidth]{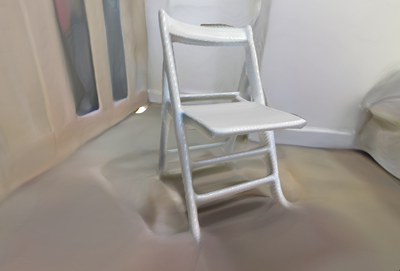} \hfill
    \includegraphics[width=0.203\linewidth]{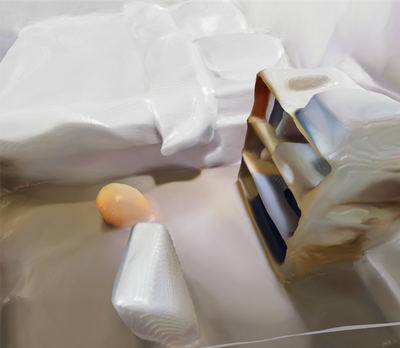} \hfill
    \includegraphics[width=0.26\linewidth]{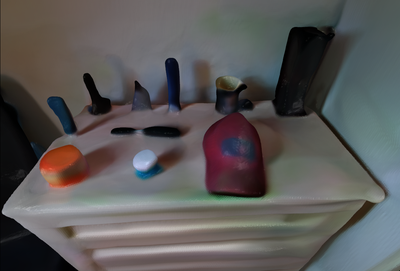} \hfill
    \includegraphics[width=0.26\linewidth]{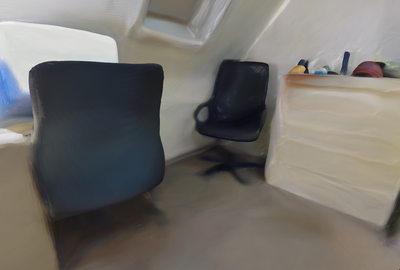} 
    \\
     \makecell{TSDF \\ Fusion} &
    \begin{subfigure}[b]{0.26\linewidth}
      \centering
      \includegraphics[width=\linewidth]{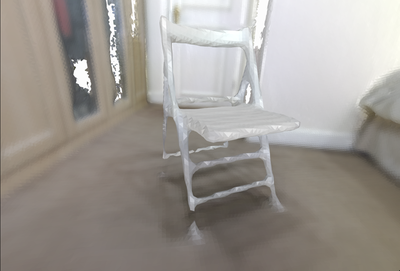} 
      \caption{Chair}
      \label{subfig:chair}
   \end{subfigure}\hfill
      \begin{subfigure}[b]{0.203\linewidth}
      \centering
      \includegraphics[width=\linewidth]{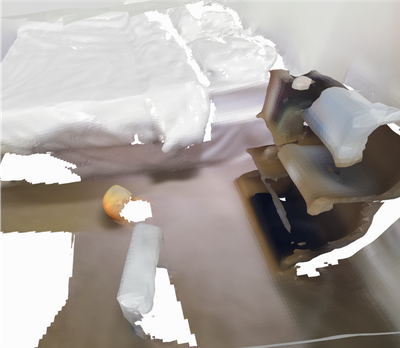} 
      \caption{Back of Objects}
       \label{subfig:back_objs}
   \end{subfigure}\hfill
    \begin{subfigure}[b]{0.26\linewidth}
      \centering
      \includegraphics[width=\linewidth]{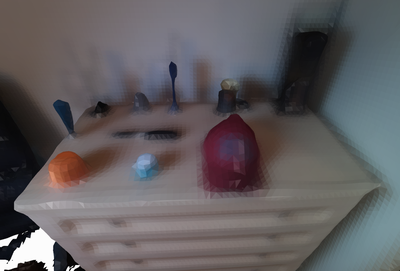} 
      \caption{Small Objects}
       \label{subfig:small_objs}
   \end{subfigure}\hfill
    \begin{subfigure}[b]{0.26\linewidth}
      \centering
      \includegraphics[width=\linewidth]{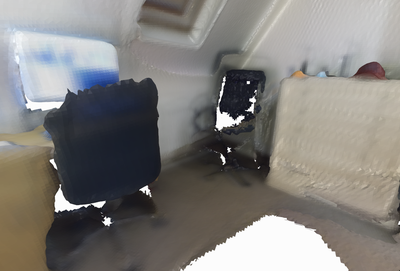}
      \caption{Black Chair}
       \label{subfig:black_chair}
   \end{subfigure}\\
  \end{tabular}
  \caption{Comparative reconstruction results in  various real scenes mapped with an Azure Kinect. White holes in the TDSF fusion results are plausibly filled in by iMAP.}
  \label{fig:real_comp}
\end{figure*}

Memory consumption for iMAP and TSDF fusion with different configuration settings is shown in Table \ref{tab:memory}. With default values of $256^3$ voxel resolution in TSDF fusion and $256$ network width in iMAP, our system can represent scenes with a factor of 60 less memory usage while obtaining similar reconstruction accuracy as seen in Table \ref{tab:replica}.

\begin{table}[htb]
  \centering
  \small
  \setlength{\tabcolsep}{0.3em}
    \begin{tabular}{lccc}
      \toprule
       \multirow{2}{*}{\textbf{iMAP} [MB] }  & Width = 128 & Width = 256  & Width = 512 \\
     & 0.26 & 1.04 & 4.19 \\
     \midrule
     \multirow{2}{*}{\textbf{TSDF Fusion} [MB] }  & Res. = 128 & Res. = 256  & Res. = 512 \\
     & 8.38 & 67.10 & 536.87 \\
      \bottomrule
    \end{tabular}%
    \caption{Memory consumption: for iMAP as a function of network size, and for TSDF fusion of voxel resolution.}
    \label{tab:memory}
        \vspace{2mm} \hrule
\end{table}

When using a real camera, in addition to better completion our method outperforms TSDF fusion in places where a depth camera does not give accurate readings as is common for black objects (Fig. \ref{subfig:black_chair}), and reflective or transparent surfaces (Fig. \ref{fig:room_comp}). This performance can be attributed to the photometric loss for reconstruction combined with the interpolation capacity of the map network.

\subsection{TUM Evaluation}
\begin{figure}[htb]
   \centering 
   \includegraphics[width=0.47\linewidth]{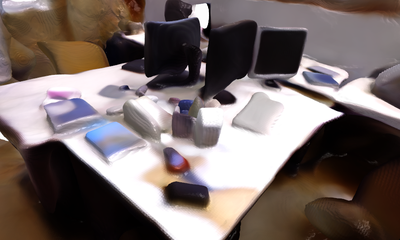}\hfill
   \includegraphics[width=0.52\linewidth]{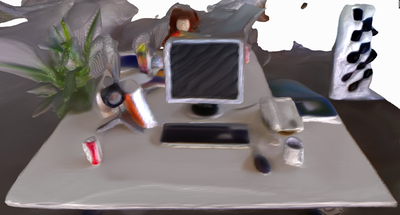}
   \includegraphics[width=\linewidth]{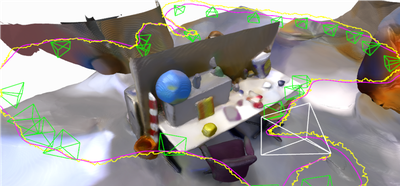}
\caption{iMAP reconstruction results for TUM dataset.}
\label{fig:tum_rec}
    \vspace{2mm} \hrule
\end{figure}

\begin{figure}[htb]
   \centering 
   \includegraphics[width=\linewidth]{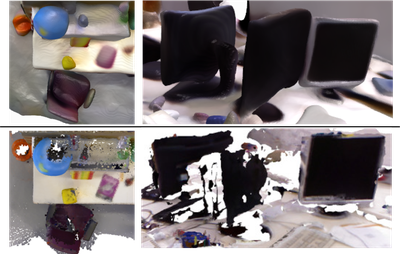}
\caption{Hole filling capacity of iMAP (top) against BAD-SLAM (bottom).}
\label{fig:tum_hole}
    \vspace{2mm} \hrule
\end{figure}

We run iMAP on three sequences from TUM RGB-D. Tracking ATE RMSE is shown in Table \ref{tab:track}. We compare with surfel-based BAD-SLAM \cite{schops2019bad}, TSDF fusion Kintinuous \cite{Whelan:etal:RSSRGBD2012}, and sparse ORB-SLAM2 \cite{Mur-Artal:etal:TRO2017}, state-of-the-art SLAM systems. In pose accuracy, iMAP does not outperform them, but is competitive with errors between 2 and 6 cm. Mesh reconstructions are shown in Figure \ref{fig:tum_rec}. In Figure \ref{fig:tum_hole} we highlight how iMAP fills in holes in unobserved regions unlike BAD-SLAM. 

\begin{table}[ht!]
  \centering
  \footnotesize
  \setlength{\tabcolsep}{0.9em}
    \begin{tabular}{lccc}
      \toprule
         & fr1/desk (cm) &  fr2/xyz (cm) &  fr3/office (cm) \\
         \midrule
        {\bf iMAP} & 4.9 & 2.0 & 5.8  \\
        {\bf BAD-SLAM} & 1.7  & 1.1  & 1.73 \\
        {\bf Kintinuous} & 3.7  &  2.9  & 3.0 \\
        {\bf ORB-SLAM2} & 1.6  & 0.4  & 1.0 \\
       \bottomrule
    \end{tabular}%
    \caption{ATE RMSE in cm on TUM RGB-D dataset.}
    \label{tab:track}
    \vspace{2mm} \hrule
\end{table}

\subsection{Ablative Analysis}
\label{subsec:ablative}

\begin{figure}[b!]
  \centering
  \includegraphics[width=0.475\linewidth]{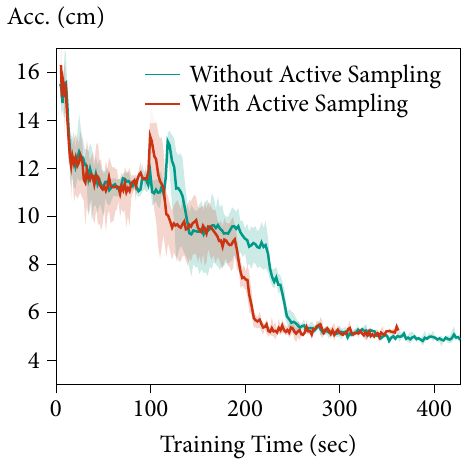}\hfill
  \includegraphics[width=0.49\linewidth]{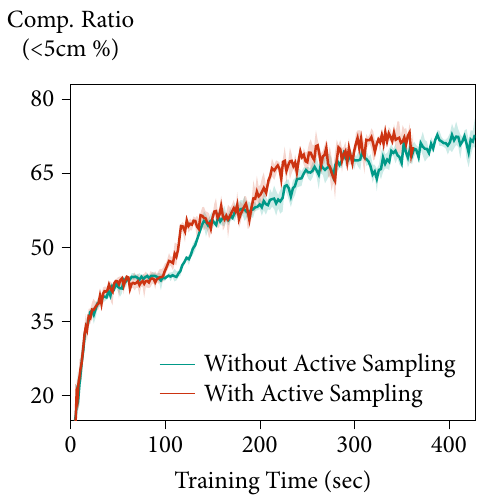}%
\caption{Active sampling obtains better completion with faster accuracy convergence than pure random sampling.}
    \vspace{2mm} \hrule
    \label{fig:active_sampling}
\end{figure}

We analyse the design choices that affect our system using the largest Replica scene: {\tt office-2} with three different random seeds. Completion ratio results and timings are shown in Table \ref{tab:timings}.  We found that network {\it width} $ = 256$, keyframe {\it window} size limit of $W=5$, and $200$ {\it pixels} samples per frame offered the best trade-off of convergence speed and accuracy. We further show in Fig. \ref{fig:active_sampling} that active sampling enables faster accuracy convergence and higher scene completion than random sampling. 

\begin{table}[ht!]
  \centering
  \footnotesize
  \setlength{\tabcolsep}{0.45em}
    \begin{tabular}{lccccccc}
      \toprule
     & \multirow{2}[2]{*}{Default} &  \multicolumn{2}{c}{Width} &  \multicolumn{2}{c}{Window}  &  \multicolumn{2}{c}{Pixels}  \\
       \cmidrule(l{2pt}r{2pt}){3-4} \cmidrule(l{2pt}r{2pt}){5-6} \cmidrule(l{2pt}r{2pt}){7-8}
      & & 128 & 512 & 3 & 10 & 100 & 400 \\
       \midrule
       \makecell[l]{{\bf Tracking} \\ {\bf Time} {[ms]} } & 101 & 80 & 173 & 84 & 144 & 74 & 160  \\
       \midrule
       \makecell[l]{{\bf Joint Optim.} \\ {\bf Time} {[ms]} } & 448 & 357 & 777 & 373 & 647 & 340 & 716 \\
       \midrule
       \makecell[l]{{\bf Comp. Ratio} \\ {[$<$5cm \%]} } & 77.22 & 75.79 & 76.91 &  75.82 & 77.35 & 77.33 & 77.49\\
       \bottomrule
    \end{tabular}%
    \caption{Timing results for tracking (6 iterations) and mapping (10 iterations), running concurrently on the same GPU. Default configuration: network width $256$, window size $5$, and $200$ samples per keyframe. Last row: completion ratio for Replica {\tt office-2}.}
    \label{tab:timings}
    \vspace{2mm} 
\end{table}

These design choices enable our online implicit SLAM system to run at 10 Hz for tracking and 2 Hz for mapping. Our experiments demonstrate the power of randomised sampling in optimisation, and highlight the key finding that it is better to iterate fast with randomly changing information than to use dense and slow iterations.

Combining geometric and photometric losses enables our system to obtain full room scale reconstructions from few keyframes; 13 on average for the 8 Replica scenes in Table \ref{tab:replica}. Using more keyframes does little to further improve scene completion as shown in Table \ref{tab:threshold}. 

\begin{table}[t!]
  \centering
  \footnotesize
  \setlength{\tabcolsep}{0.2em}
    \begin{tabular}{lcccc}
      \toprule
         & $t_P=0.55$ &  $t_P=0.65$ &  $t_P=0.75$ &  $t_P=0.85$ \\
         \midrule
        {\bf \# Keyframes} & 8 & 10 & 14 & 24  \\
        {\bf Comp. Ratio} [$<$5cm \%] & 74.11  & 77.22  & 76.84  & 78.03 \\
       \bottomrule
    \end{tabular}%
    \caption{Number of keyframe and completion ratio results for different selection thresholds in Replica {\tt office-2}.}
    \label{tab:threshold}
    \vspace{2mm} \hrule
\end{table}

\begin{figure}[t]
   \centering
   \includegraphics[width=\linewidth]{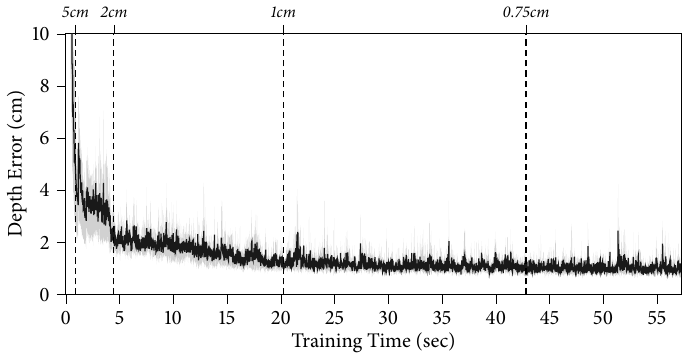}
\caption{Reaching 5cm, 2cm, 1cm and 0.75cm depth error requires around 1, 4, 20, 43 seconds respectively.}
\label{fig:time}
    \vspace{2mm} \hrule
    \vspace{-1em}
\end{figure}

\begin{figure}[t]
    \centering
   \includegraphics[width=\linewidth]{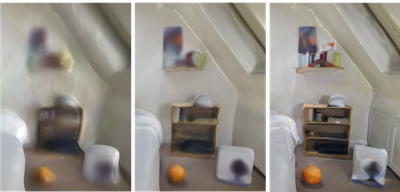}
\caption{Evolution of reconstruction detail.}
\label{fig:detail}
    \vspace{2mm} \hrule
    \vspace{-1em}
\end{figure}

Implicit scene networks have the property of converging fast to low frequency shapes before adding higher frequency scene details. Fig. \ref{fig:time} shows network training from a static camera averaged over 5 different real scenes. The depth loss falls below 5cm in under a second; under 2cm in 4 seconds; then continues to decrease slowly. When mapping a new scene our system takes seconds to get a coarse reconstruction and minutes to add in fine details. In Fig. \ref{fig:detail} we show how the system starts with a rough reconstruction and adds detail as the network trains and the camera moves closer to objects. This is a useful property in SLAM as it enables live tracking to work even when moving to unexplored regions.

\section{Conclusions}

We pose dense SLAM as real-time continual learning and show that an MLP can be trained from scratch as the only scene representation in a live system, thus enabling an RGB-D camera to construct and track against a complete and accurate volumetric model of room-scale scenes. 
The keys to the real-time but long-term SLAM performance of our method are: parallel tracking and mapping, loss-guided pixel sampling for rapid optimisation, and intelligent keyframe selection as replay to avoid network forgetting. Future directions for iMAP include how to make more structured and compositional representations that reason explicitly about the self similarity in scenes.

\section*{Acknowledgements}
Research presented here has been supported by Dyson Technology Ltd. We thank Kentaro Wada, Tristan Laidlow, and Shuaifeng Zhi for fruitful discussions.

{\small
\bibliographystyle{ieee_fullname}
\bibliography{robotvision.bib}

\begin{thebibliography}{10}\itemsep=-1pt

\bibitem{Bloesch:etal:CVPR2018}
M. Bloesch, J. Czarnowski, R. Clark, S. Leutenegger, and A.~J. Davison.
\newblock {CodeSLAM} --- learning a compact, optimisable representation for
  dense visual {SLAM}.
\newblock In {\em {Proceedings of the {IEEE} Conference on Computer Vision and
  Pattern Recognition ({CVPR})}}, 2018.

\bibitem{chabra2020deep}
Rohan Chabra, Jan~Eric Lenssen, Eddy Ilg, Tanner Schmidt, Julian Straub, Steven
  Lovegrove, and Richard Newcombe.
\newblock {Deep Local Shapes}: Learning local {SDF} priors for detailed 3d
  reconstruction.
\newblock {\em {Proceedings of the European Conference on Computer Vision
  ({ECCV})}}, 2020.

\bibitem{chibane2020neural}
Julian Chibane, Gerard Pons-Moll, et~al.
\newblock Neural unsigned distance fields for implicit function learning.
\newblock {\em {Neural Information Processing Systems ({NIPS})}}, 2020.

\bibitem{Curless:Levoy:SIGGRAPH1996}
B. Curless and M. Levoy.
\newblock {A volumetric method for building complex models from range images}.
\newblock In {\em {Proceedings of {SIGGRAPH}}}, 1996.

\bibitem{dai2020sg}
Angela Dai, Christian Diller, and Matthias Nie{\ss}ner.
\newblock {SG-NN}: Sparse generative neural networks for self-supervised scene
  completion of {RGB-D} scans.
\newblock In {\em {Proceedings of the {IEEE} Conference on Computer Vision and
  Pattern Recognition ({CVPR})}}, 2020.

\bibitem{Dai:etal:ACMTOG2017}
Angela Dai, Matthias Nie{\ss}ner, Michael Zollh\"{o}fer, Shahram Izadi, and
  Christian Theobalt.
\newblock {BundleFusion: Real-time Globally Consistent 3D Reconstruction using
  On-the-fly Surface Re-integration}.
\newblock {\em {{{ACM} Transactions on Graphics ({TOG})}}}, 36(3):24:1--24:18,
  2017.

\bibitem{grossberg1982does}
Stephen Grossberg.
\newblock How does a brain build a cognitive code?
\newblock In {\em Studies of mind and brain}, pages 1--52. Springer, 1982.

\bibitem{Keller:etal:3DV2013}
M. Keller, D. Lefloch, M. Lambers, S. Izadi, T. Weyrich, and A. Kolb.
\newblock {Real-time 3D Reconstruction in Dynamic Scenes using Point-based
  Fusion}.
\newblock In {\em Proc. of Joint 3DIM/3DPVT Conference (3DV)}, 2013.

\bibitem{Kingma:etal:ICLR2015}
Diederik~P. Kingma and Jimmy Ba.
\newblock {ADAM}: A method for stochastic optimization.
\newblock In {\em {Proceedings of the International Conference on Learning
  Representations ({ICLR})}}, 2015.

\bibitem{kirkpatrick2017overcoming}
James Kirkpatrick, Razvan Pascanu, Neil Rabinowitz, Joel Veness, Guillaume
  Desjardins, Andrei~A Rusu, Kieran Milan, John Quan, Tiago Ramalho, Agnieszka
  Grabska-Barwinska, et~al.
\newblock Overcoming catastrophic forgetting in neural networks.
\newblock {\em Proceedings of the national academy of sciences},
  114(13):3521--3526, 2017.

\bibitem{Klein:Murray:ISMAR2007}
G. Klein and D.~W. Murray.
\newblock {Parallel Tracking and Mapping for Small {AR} Workspaces}.
\newblock In {\em {Proceedings of the International Symposium on Mixed and
  Augmented Reality ({ISMAR})}}, 2007.

\bibitem{lesort2019generative}
Timoth{\'e}e Lesort, Hugo Caselles-Dupr{\'e}, Michael Garcia-Ortiz, Andrei
  Stoian, and David Filliat.
\newblock Generative models from the perspective of continual learning.
\newblock In {\em 2019 International Joint Conference on Neural Networks
  (IJCNN)}, pages 1--8. IEEE, 2019.

\bibitem{maltoni2019continuous}
Davide Maltoni and Vincenzo Lomonaco.
\newblock Continuous learning in single-incremental-task scenarios.
\newblock {\em Neural Networks}, 116:56--73, 2019.

\bibitem{mescheder2019occupancy}
Lars Mescheder, Michael Oechsle, Michael Niemeyer, Sebastian Nowozin, and
  Andreas Geiger.
\newblock {Occupancy Networks}: Learning {3D} reconstruction in function space.
\newblock In {\em {Proceedings of the {IEEE} Conference on Computer Vision and
  Pattern Recognition ({CVPR})}}, 2019.

\bibitem{Mildenhall:etal:ECCV2020}
Ben Mildenhall, Pratul~P. Srinivasan, Matthew Tancik, Jonathan~T. Barron, Ravi
  Ramamoorthi, and Ren Ng.
\newblock {NeRF}: Representing scenes as neural radiance fields for view
  synthesis.
\newblock In {\em {Proceedings of the European Conference on Computer Vision
  ({ECCV})}}, 2020.

\bibitem{Mur-Artal:etal:TRO2017}
R. Mur-Artal and J.~D. Tard{\'o}s.
\newblock {ORB-SLAM2: An Open-Source SLAM System for Monocular, Stereo, and
  RGB-D Cameras}.
\newblock {\em {{IEEE} Transactions on Robotics ({T-RO})}}, 33(5):1255--1262,
  2017.

\bibitem{Newcombe:etal:ISMAR2011}
R.~A. Newcombe, S. Izadi, O. Hilliges, D. Molyneaux, D. Kim, A.~J. Davison, P.
  Kohli, J. Shotton, S. Hodges, and A. Fitzgibbon.
\newblock {{KinectFusion}: Real-Time Dense Surface Mapping and Tracking}.
\newblock In {\em {Proceedings of the International Symposium on Mixed and
  Augmented Reality ({ISMAR})}}, 2011.

\bibitem{Newcombe:etal:ICCV2011}
R.~A. Newcombe, S. Lovegrove, and A.~J. Davison.
\newblock {{DTAM}: Dense Tracking and Mapping in Real-Time}.
\newblock In {\em {Proceedings of the International Conference on Computer
  Vision ({ICCV})}}, 2011.

\bibitem{park2019deepsdf}
Jeong~Joon Park, Peter Florence, Julian Straub, Richard Newcombe, and Steven
  Lovegrove.
\newblock {DeepSDF}: Learning continuous signed distance functions for shape
  representation.
\newblock In {\em {Proceedings of the {IEEE} Conference on Computer Vision and
  Pattern Recognition ({CVPR})}}, 2019.

\bibitem{paszke2019pytorch}
Adam Paszke, Sam Gross, Francisco Massa, Adam Lerer, James Bradbury, Gregory
  Chanan, Trevor Killeen, Zeming Lin, Natalia Gimelshein, Luca Antiga, et~al.
\newblock {PyTorch}: An imperative style, high-performance deep learning
  library.
\newblock In {\em {Neural Information Processing Systems ({NIPS})}}, 2019.

\bibitem{Peng2020ECCV}
Songyou Peng, Michael Niemeyer, Lars Mescheder, Marc Pollefeys, and Andreas
  Geiger.
\newblock Convolutional occupancy networks.
\newblock In {\em {Proceedings of the European Conference on Computer Vision
  ({ECCV})}}, 2020.

\bibitem{rolnick2019experience}
David Rolnick, Arun Ahuja, Jonathan Schwarz, Timothy Lillicrap, and Gregory
  Wayne.
\newblock Experience replay for continual learning.
\newblock In {\em {Neural Information Processing Systems ({NIPS})}}, 2019.

\bibitem{rusu2016progressive}
Andrei~A Rusu, Neil~C Rabinowitz, Guillaume Desjardins, Hubert Soyer, James
  Kirkpatrick, Koray Kavukcuoglu, Razvan Pascanu, and Raia Hadsell.
\newblock Progressive neural networks.
\newblock {\em arXiv preprint arXiv:1606.04671}, 2016.

\bibitem{schops2019bad}
Thomas Schops, Torsten Sattler, and Marc Pollefeys.
\newblock {BAD SLAM}: Bundle adjusted direct {RGB-D} {SLAM}.
\newblock In {\em {Proceedings of the {IEEE} Conference on Computer Vision and
  Pattern Recognition ({CVPR})}}, 2019.

\bibitem{schwarz2018progress}
Jonathan Schwarz, Jelena Luketina, Wojciech~M Czarnecki, Agnieszka
  Grabska-Barwinska, Yee~Whye Teh, Razvan Pascanu, and Raia Hadsell.
\newblock Progress \& compress: A scalable framework for continual learning.
\newblock {\em arXiv preprint arXiv:1805.06370}, 2018.

\bibitem{shin2017continual}
Hanul Shin, Jung~Kwon Lee, Jaehong Kim, and Jiwon Kim.
\newblock Continual learning with deep generative replay.
\newblock In {\em {Neural Information Processing Systems ({NIPS})}}, 2017.

\bibitem{Sitzmann:etal:NIPS2020}
Vincent Sitzmann, Julien Martel, Alexander Bergman, David Lindell, and Gordon
  Wetzstein.
\newblock Implicit neural representations with periodic activation functions.
\newblock {\em {Neural Information Processing Systems ({NIPS})}}, 2020.

\bibitem{Smith:Cheeseman:IJRR1986}
R.~C. Smith and P. Cheeseman.
\newblock {On the Representation and Estimation of Spatial Uncertainty}.
\newblock {\em {International Journal of Robotics Research ({IJRR})}},
  5(4):56--68, Dec. 1986.

\bibitem{Straub:etal:ARXIV2019}
Julian Straub, Thomas Whelan, Lingni Ma, Yufan Chen, Erik Wijmans, Simon Green,
  Jakob~J Engel, Raul Mur-Artal, Carl Ren, Shobhit Verma, et~al.
\newblock {The Replica Dataset}: A digital replica of indoor spaces.
\newblock {\em arXiv preprint arXiv:1906.05797}, 2019.

\bibitem{Sturm:etal:IROS2012}
J. Sturm, N. Engelhard, F. Endres, W. Burgard, and D. Cremers.
\newblock {A Benchmark for the Evaluation of {RGB-D} {SLAM} Systems}.
\newblock In {\em {Proceedings of the {IEEE/RSJ} Conference on Intelligent
  Robots and Systems ({IROS})}}, 2012.

\bibitem{Sucar:etal:3DV2020}
Edgar Sucar, Kentaro Wada, and Andrew Davison.
\newblock {NodeSLAM}: Neural object descriptors for multi-view shape
  reconstruction.
\newblock In {\em {Proceedings of the International Conference on 3D Vision
  ({3DV})}}, 2020.

\bibitem{Tancik:etal:NIPS2020}
Matthew Tancik, Pratul Srinivasan, Ben Mildenhall, Sara Fridovich-Keil, Nithin
  Raghavan, Utkarsh Singhal, Ravi Ramamoorthi, Jonathan Barron, and Ren Ng.
\newblock Fourier features let networks learn high frequency functions in low
  dimensional domains.
\newblock {\em {Neural Information Processing Systems ({NIPS})}}, 2020.

\bibitem{tang2020deep}
Danhang Tang, Saurabh Singh, Philip~A Chou, Christian Hane, Mingsong Dou, Sean
  Fanello, Jonathan Taylor, Philip Davidson, Onur~G Guleryuz, Yinda Zhang,
  et~al.
\newblock Deep implicit volume compression.
\newblock In {\em {Proceedings of the {IEEE} Conference on Computer Vision and
  Pattern Recognition ({CVPR})}}, 2020.

\bibitem{Vespa:etal:RAL2018}
Emanuele Vespa, Nikolay Nikolov, Marius Grimm, Luigi Nardi, Paul~HJ Kelly, and
  Stefan Leutenegger.
\newblock Efficient octree-based volumetric {SLAM} supporting signed-distance
  and occupancy mapping.
\newblock {\em {{IEEE} Robotics and Automation Letters}}, 2018.

\bibitem{wang2021nerf}
Zirui Wang, Shangzhe Wu, Weidi Xie, Min Chen, and Victor~Adrian Prisacariu.
\newblock {NeRF--}: Neural radiance fields without known camera parameters.
\newblock {\em arXiv preprint arXiv:2102.07064}, 2021.

\bibitem{weder2020routedfusion}
Silvan Weder, Johannes Schonberger, Marc Pollefeys, and Martin~R Oswald.
\newblock {RoutedFusion}: Learning real-time depth map fusion.
\newblock In {\em {Proceedings of the {IEEE} Conference on Computer Vision and
  Pattern Recognition ({CVPR})}}, 2020.

\bibitem{Whelan:etal:RSS2015}
T. Whelan, S. Leutenegger, R.~F. Salas-Moreno, B. Glocker, and A.~J. Davison.
\newblock {ElasticFusion}: Dense {SLAM} without a pose graph.
\newblock In {\em {Proceedings of Robotics: Science and Systems ({RSS})}},
  2015.

\bibitem{Whelan:etal:RSSRGBD2012}
T. Whelan, J.~B. McDonald, M. Kaess, M. Fallon, H. Johannsson, and J.~J.
  Leonard.
\newblock {Kintinuous: Spatially Extended KinectFusion}.
\newblock In {\em {Workshop on {RGB-D}: Advanced Reasoning with Depth Cameras,
  in conjunction with Robotics: Science and Systems}}, 2012.

\bibitem{yen2020inerf}
Lin Yen-Chen, Pete Florence, Jonathan~T Barron, Alberto Rodriguez, Phillip
  Isola, and Tsung-Yi Lin.
\newblock {iNeRF}: Inverting neural radiance fields for pose estimation.
\newblock {\em arXiv preprint arXiv:2012.05877}, 2020.

\end{thebibliography}
}

\end{document}